\documentclass[runningheads]{llncs}
\usepackage{graphicx}

\usepackage{tikz}
\usepackage{comment} 
\usepackage{amsmath,amssymb} 
\usepackage{color}

\usepackage{xcolor}
\usepackage{multirow}
\usepackage{makecell}
\usepackage{subcaption}
\begin{document}
\pagestyle{headings}
\mainmatter
\def\ECCVSubNumber{6051}  

\title{VLANet: Video-Language Alignment Network for Weakly-Supervised Video Moment Retrieval} 

\titlerunning{Video-Language Alignment Network}
%
\newcommand*\samethanks[1][\value{footnote}]{\footnotemark[#1]}

\author{Minuk Ma\thanks{Both authors have equally contributed.} \and
Sunjae Yoon\samethanks \and
Junyeong Kim \and
Youngjoon Lee \and
Sunghun Kang \and
Chang D. Yoo}

%
\authorrunning{Ma, Yoon, Kim, Lee, Kang, Yoo}
%

\institute{Korea Advanced Institute of Science and Technology, Daejeon, Republic of Korea
\email{\{akalsdnr,dbstjswo505,junyeong.kim,yjlee22,sunghun.kang,cd\_yoo\}@kaist.ac.kr}}
\maketitle
\begin{abstract}
Video Moment Retrieval (VMR) is a task to localize the temporal moment in untrimmed video specified by natural language query. 
For VMR, several methods that require full supervision for training have been proposed.
Unfortunately, acquiring a large number of training videos with labeled temporal boundaries for each query is a labor-intensive process. 
This paper explores a method for performing VMR in a weakly-supervised manner (wVMR): training is performed without temporal moment labels but only with the text query that describes a segment of the video. 
Existing methods on wVMR generate multi-scale proposals and apply query-guided attention mechanism to highlight the most relevant proposal. 
To leverage the weak supervision, contrastive learning is used which predicts higher scores for the correct video-query pairs than for the incorrect pairs. 
It has been observed that a large number of candidate proposals, coarse query representation, and one-way attention mechanism lead to blurry attention map which limits the localization performance.
To address this issue, Video-Language Alignment Network (VLANet) is proposed that learns a sharper attention by pruning out spurious candidate proposals and applying a multi-directional attention mechanism with fine-grained query representation.
The Surrogate Proposal Selection module selects a proposal based on the proximity to the query in the joint embedding space, and thus substantially reduces candidate proposals which leads to lower computation load and sharper attention.
Next, the Cascaded Cross-modal Attention module considers dense feature interactions and multi-directional attention flows to learn the multi-modal alignment. 
VLANet is trained end-to-end using contrastive loss which enforces semantically similar videos and queries to cluster. 
The experiments show that the method achieves state-of-the-art performance on Charades-STA and DiDeMo datasets. 
\keywords{Multi-modal learning, weakly-supervised learning, video moment retrieval}
\end{abstract}

\section{Introduction}
\begin{figure}[t]
	\centering
	\includegraphics[width=11cm,height=5.55cm]{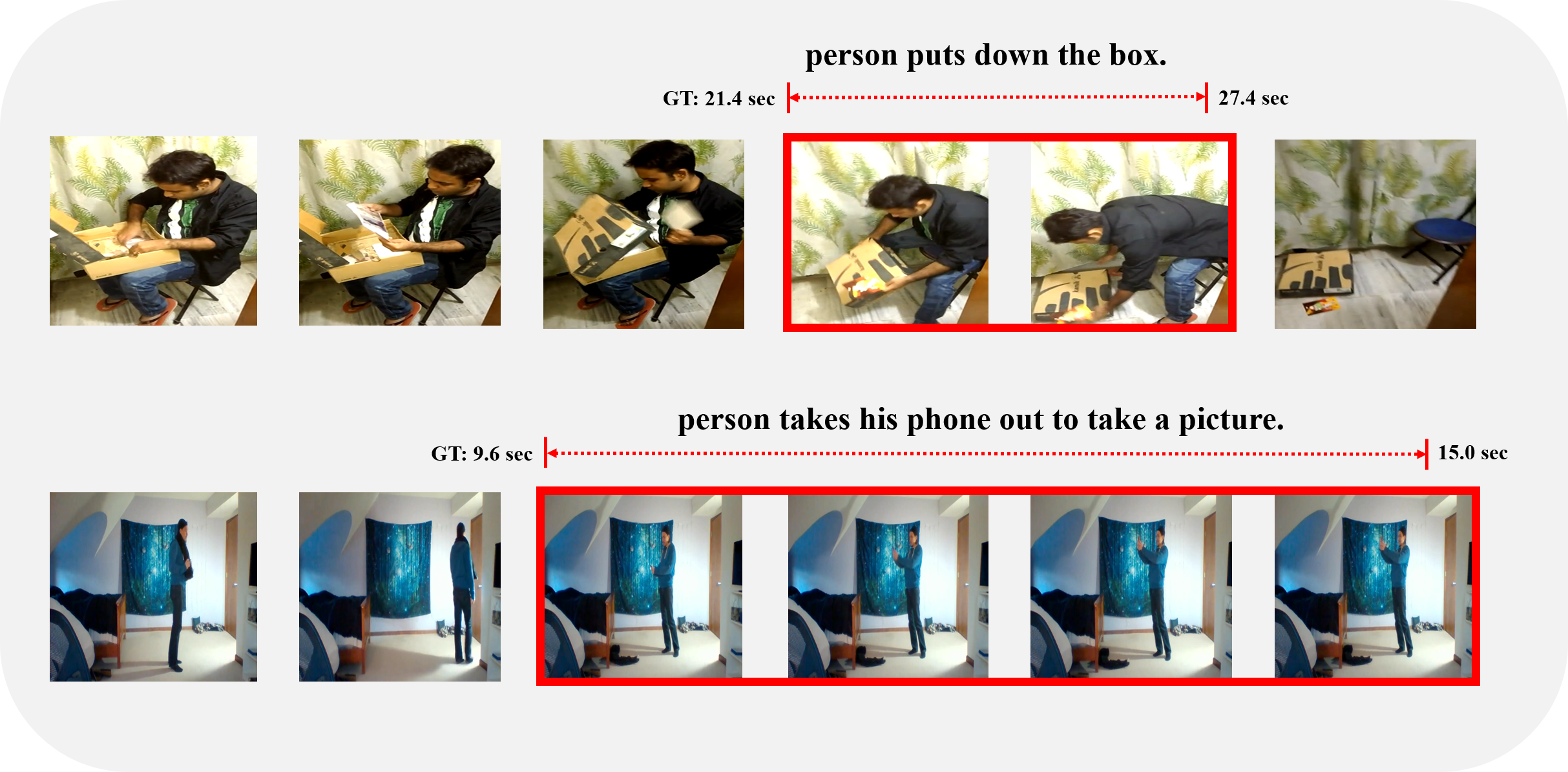}
	\caption{Illustration of video moment retrieval task. The goal is to search the temporal boundary of the video moment that is most relevant to the given natural language query.}
	\label{fig:task}
\end{figure}

Video moment retrieval (VMR) is a task to find a temporal moment in untrimmed video specified by a text description as illustrated in Figure \ref{fig:task}. With the rising number of videos along with the need for a more detailed and refined search capability that demand a better understanding of the video, the task of Video Moment Retrieval is drawing appreciable attention.

A number of fully-supervised methods that learn from a set of videos with ground-truth time stamps corresponding to a given query have been proposed \cite{TALL,MCN,EFRC,MAN}.  
For these methods, a large-scale video dataset that requires the laborious burden of temporally annotating the boundaries corresponding to each query is a sine qua non.
In general, the performance of a fully-supervised method hinges on the quality of the dataset; however, for VMR, temporal boundaries are often ambiguous to annotate and may act as noise in the learning process. 

Recently, weakly-supervised VMR (wVMR) \cite{that,SCN} that does not require the temporal boundary annotation for each query has been studied.
To leverage the weak supervision, contrastive learning is applied such that higher scores are predicted for the correct video-query pairs than for incorrect pairs. 
This learning process improves the accuracy of the attention mechanism which plays a vital role in wVMR.
Inspired by recent methods \cite{that,SCN}, this paper addresses two critical challenges: (1) generating appropriate multi-scale video candidate proposals, and (2) learning the latent alignment between the text query and the retrieved video segment.  

The first challenge is that the video segment proposals should be adequate in number to give high recall without excessive computational load, and the video segment should be of appropriate length to have high intersection-of-union (IoU) with ground truth. 
Previous methods \cite{TALL,MCN,that,SCN} greedily generated video candidate proposals using a pre-defined set of multi-scale sliding windows. 
As a consequence, these methods generally produce large number of multi-scale proposals which increase the chance of achieving high recall at the expense of high computational cost. 
When an attention mechanism is used thereafter to weigh the proposals, the attention becomes blurry as there are too many proposals to attend.

The second challenge is to learn a similarity measure between video segment and text query without ground truth annotation. 
In \cite{that}, a text-to-video attention mechanism is incorporated to learn the joint embedding space of video and text query. 
More accurate multi-modal similarity could be attained with a text query representation that is more effective in interacting with video frame feature.
Representing the text query as the last hidden feature of the Gated Recurrent Unit (GRU), as used in some previous methods \cite{that,SCN}, is overly simplistic .
In addition, applying one-way attention from query to video is not sufficient to bring out the most prominent feature in the video and query. 
Recent studies in Visual Question Answering \cite{MCAN,InterIntra,SMAN} have explored the possibility of applying multi-directional attention flows that include both inter- and intra-modality attention.
This paper devises an analogous idea for the problem of wVMR, and validate its effectiveness in retrieving the moment using the weak labels. 

To rise to the challenge, this paper proposes a Video-Language Alignment Network (VLANet) for weakly-supervised video moment retrieval. 
As a first step, the word-level query representation is obtained by stacking all intermediate hidden features of GRU. 
Video is divided into overlapping multi-scale segment groups where the segments within each group share a common starting time. 
Then, the Surrogate Proposal Selection module selects one surrogate from each group which reduces the number of effective proposals for more accurate attention. 
To consider the multi-directional interactions between each surrogate proposal and query, the Cascaded Cross-modal Attention (CCA) module performs both intra- and inter-modality attention. 
The CCA module performs self-attention on each modality: video to video (V2V) and query to query (Q2Q), which considers the intra-modal relationships. 
Thereafter, the CCA module performs cross-modal attention from query to video (Q2V), video to query (V2Q) and finally attended query to attended video (Q2V). 
This cross-modal attention considers the inter-modal relationships that is critical in learning the multi-modal alignment. 
To leverage the weak labels of video-query pairs, VLANet is trained in an end-to-end manner using contrastive loss that enforces semantically similar videos and queries to cluster in the joint embedding space. 
The experiment results show that the VLANet achieves state-of-the-art performance on Charades-STA and DiDeMo datasets.
Extensive ablation study and qualitative analyses validate the effectiveness of the proposed method and provide interpretability. 

\section{Related Work}
\subsection{Temporal Action Detection}
The goal of temporal action detection is to predict the temporal boundary and category for each action instance in untrimmed videos. 
Existing works are divided into two groups: the fully-supervised and weakly-supervised. 
Zhao \textit{et al.} \cite{SSN} proposed a structured segment network that models the temporal structure of each action instance by a structured temporal pyramid. 
Gao \textit{et al.} \cite{CBR} proposed Cascaded Boundary Regression which uses temporal coordinate regression to refine the temporal boundaries of the sliding windows. 
Lin \textit{et al.} \cite{BSN} proposed Boundary Sensitive Network that first classifies each frame as the start, middle, or end, then directly combines these boundaries as proposals. 

In the weakly-supervised settings, however, only the coarse video-level labels are available instead of the exact temporal boundaries.
Wang \textit{et al.} \cite{UntrimmedNet} proposed UntrimmedNet that couples two components, the classification module, and the selection module, to learn the action models and reason about the temporal duration of action instances, respectively.
Nguyen \textit{et al.} \cite{STPN} proposed a Sparse Temporal Pooling Network that identifies a sparse subset of key segments associated with the target actions in a video using an attention module and fuse the key segments using adaptive temporal pooling.
Shou \textit{et al.} \cite{AutoLoc} proposed AutoLoc that uses Outer-Inner-Contrastive loss to automatically discover the required segment-level supervision to train a boundary predictor.
Liu \textit{et al.} \cite{CleanNet} proposed CleanNet that leverages an additional temporal contrast constraint so that the high-evaluation-score action proposals have a higher probability to overlap with the ground truth action instances. 

\subsection{Video Moment Retrieval}
The VMR task is focused on localizing the temporal moment that is semantically aligned with the given natural language query. 
For this task, various supervised methods have been proposed \cite{TALL,MCN,EFRC,MAN}.
In Gao \textit{et al.} \cite{TALL} and Hendricks \textit{et al.} \cite{MCN}, candidate moments are sampled using sliding windows of various lengths, and multi-modal fusion is performed to estimate the correlation between the queries and video moments. 
Xu \textit{et al.} \cite{EFRC} proposed a model that integrates vision and language features using attention mechanisms and leverages video captioning as an auxiliary task. 
Zhang \textit{et al.} \cite{MAN} proposed Moment Alignment Network (MAN) that considers the relationships between proposals as a structured graph, and devised an iterative algorithm to train a revised graph convolution network.

Recently, the task was studied under the weakly-supervised setting \cite{Duan,that,SCN}. 
Duan \textit{et al.} \cite{Duan} proposed to decompose weakly-supervised dense event captioning in videos (WS-DEC) into a pair of dual problems: event captioning and sentence localization. 
They proposed a cycle system to train the model based on the assumption that each caption describes only one temporal segment. 
Mithun \textit{et al.} \cite{that} proposed Text-Guided-Attention (TGA) model that learns a joint representation between video and sentence. 
The attention weight is used to retrieve the relevant moment at test time. 
Lin \textit{et al.} \cite{SCN} proposed Semantic Completion Network (SCN) that selects the top-K proposals considering exploration and exploitation, and measures the semantic similarity between the video and query.
As an auxiliary task, SCN takes the masked sentence as input and predicts the masked words from visual representations. 

\section{Method}
\begin{figure}[t]
	\centering
	\includegraphics[width=\textwidth]{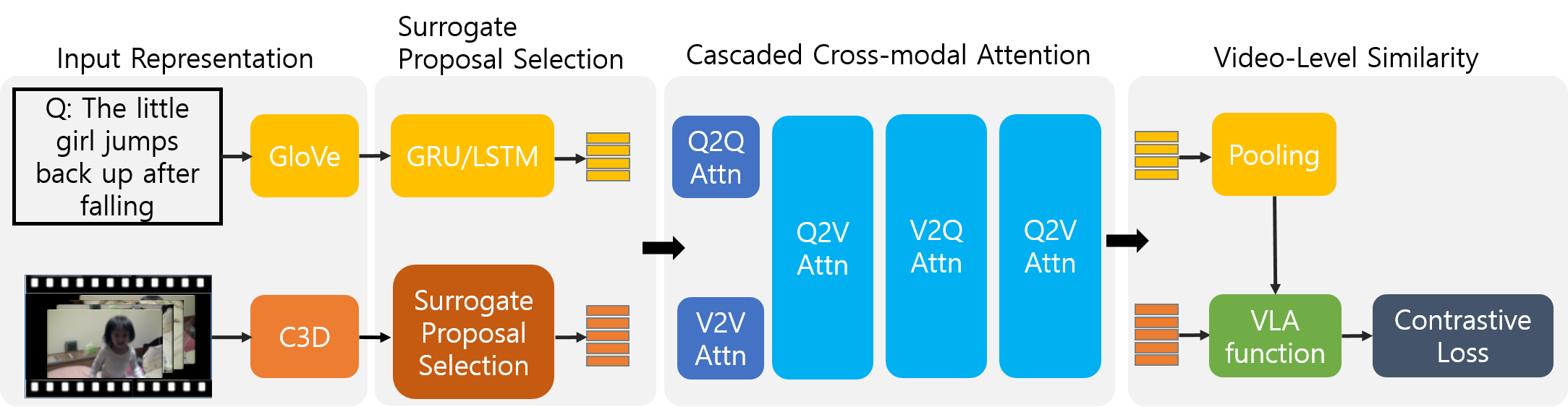}
	\caption{Illustration of VLANet architecture. The Surrogate Proposal Selection module prunes out irrelevant proposals based on the similarity metric. Cascaded Cross-modal Attention considers various attention flows to learn multi-modal alignment. The network is trained end-to-end using contrastive loss.}
	\label{fig:method}
\end{figure}

\subsection{Method Overview}
\label{ssec:method_overview}
Figure \ref{fig:method} illustrates the overall VLANet architecture. 
The input text query is embedded using GloVe \cite{GloVe} after which each embedded representation is fed into a GRU \cite{GRU}. In the meanwhile, the video is embedded based on C3D \cite{C3D}.
Video is divided into overlapping multi-scale segment groups where the proposals within each group share a common starting time. 
Given the video and query representations $V$ and $Q$, the similarity $c$ between video and query is evaluated by the Cascaded Cross-modal Attention (CCA) module. 
The learned attention weights by CCA are used to localize the relevant moment at test time.
A video-query pair $(V,Q)$ is positive if it is in the training data; otherwise, it is negative. 
The network is trained in an end-to-end manner using contrastive loss to enforce the scores of the positive pairs to be higher than those of the negative pairs. 
In practice, the negative pairs are randomly sampled in a batch.

\subsection{Input representation}
\subsubsection{Query representation}
Gated Recurrent Unit (GRU) \cite{GRU} is used for encoding the sentences. 
Each word of the query is embedded using GloVe and sequentially fed into a GRU. 
Prior methods \cite{that} use only the final hidden feature of GRU to represent the whole sentence, which leads to the loss of information by excluding the interactions between frame- and word-level features of video and query. 
Motivated by recent works in visual question answering \cite{InterIntra,MCAN}, this paper uses all intermediate hidden features of the GRU. 
The query $Q$ is represented as:
\begin{eqnarray}
    Q = \left[ {\bf w}_1 \ {\bf w}_2 \ \cdots \ {\bf w}_M \right]
\end{eqnarray}

where ${\bf w}_m \in \mathbb{R}^D$ denotes the $m$-th GRU hidden feature, and $D$ is the dimension of the hidden feature. 
Each ${\bf w}_m$ is L2 normalized to output a unit vector.

\subsubsection{Video representation}
Video is encoded using a C3D \cite{C3D} model pre-trained on Sports-1M dataset \cite{Sports1M} as in \cite{TALL}. 
The feature was extracted at every 16 frames for Charades-STA.
The VGG16 model \cite{VGG} is used for frame-level feature extraction for DiDeMo dataset following \cite{MCN}. 
Both C3D and VGG16 features were extracted from the penultimate fully-connected layer, which results in the feature dimension of 4096.

\subsubsection{Video proposal generation}

\begin{figure}[t]
        \centering
        \begin{subfigure}[t]{0.45\textwidth}
            \includegraphics[width=\linewidth, height=10em]{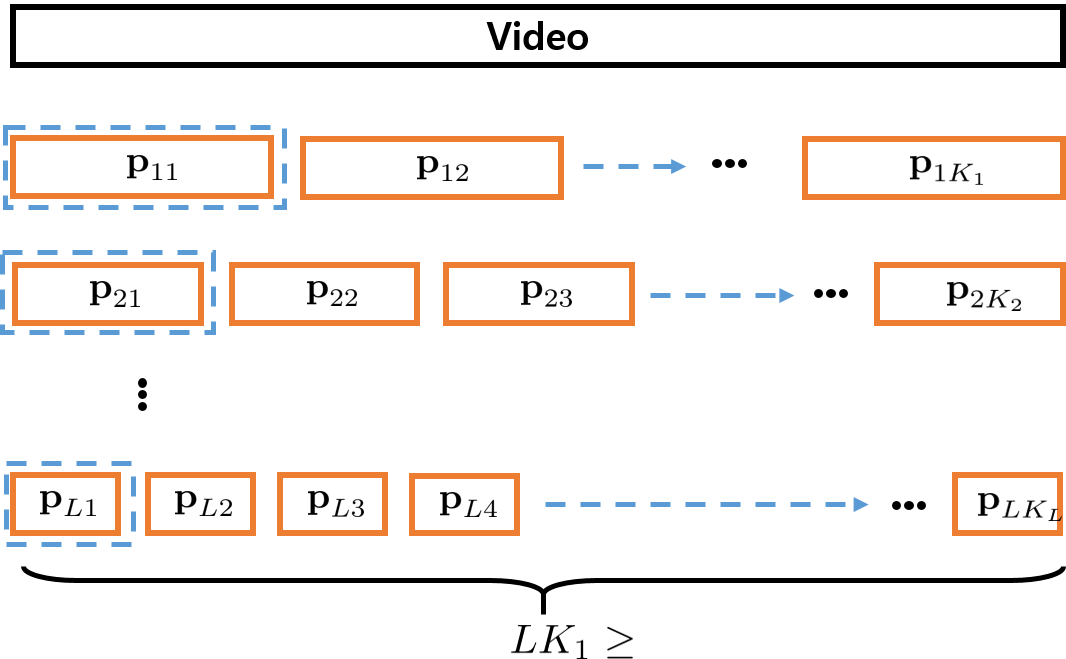}
            \caption{Multi-scale proposal generation}
        \end{subfigure}
        \begin{subfigure}[t]{0.45\textwidth}
            \includegraphics[width=\linewidth, height=10em]{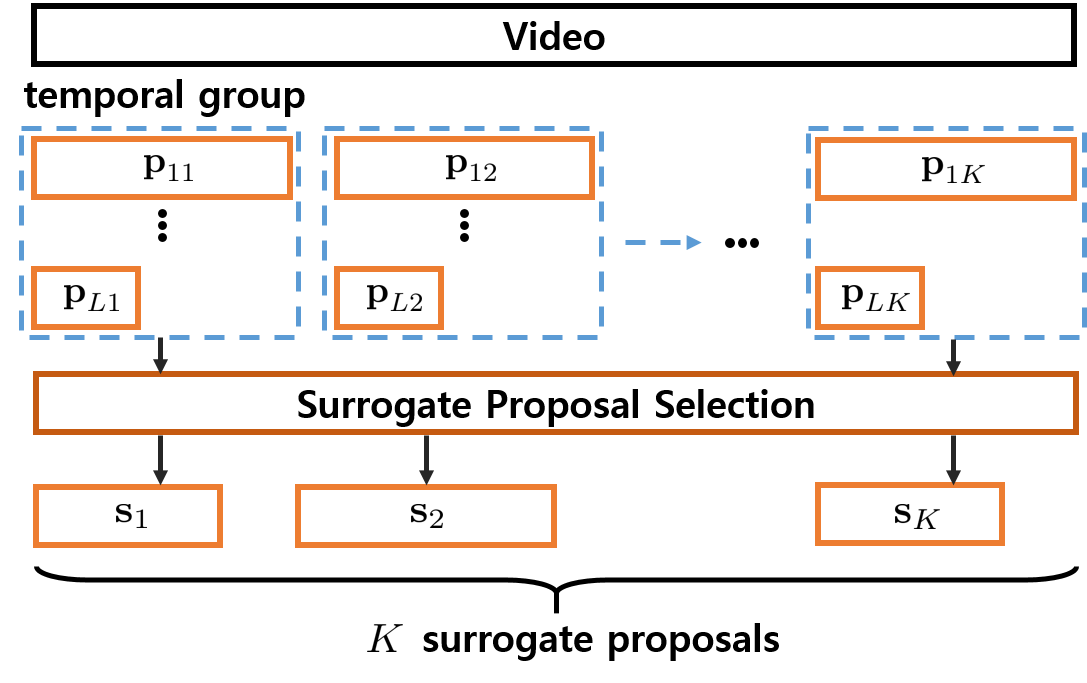}
            \caption{Surrogate Proposal Selection}
        \end{subfigure}
        \caption{Comparison between the previous and the proposed proposal generation method. (a) generates large number of proposals of various lengths. (b) groups the proposals, and selects the surrogate proposals based on the proximity to the query. }
        \label{fig:proposal}
\end{figure}

As depicted in image Figure \ref{fig:proposal}(a) previous methods \cite{that,SCN} generated proposals using multi-scale sliding windows.
Meanwhile, as in Figure \ref{fig:proposal}(b), VLANet organizes the multi-scale windows in segment groups such that within a group, all windows start at the same time instance.
Each group will have the same number of windows of fixed scales. 
The interval between the starting times of each segment group is regular. 
With $K$ segment groups and $L$ multi-scale proposals, the total number of proposals is $K \cdot L$.
Then, the video $V$ is represented by:

\begin{eqnarray}
    V = \begin{bmatrix}
        {\bf p}_{11} & {\bf p}_{12} & \cdots & {\bf p}_{1K} \\
        {\bf p}_{21} & {\bf p}_{22} & \cdots & {\bf p}_{2K} \\
        \vdots &        & \cdots &        \\
        {\bf p}_{L1} & {\bf p}_{L2} & \cdots & {\bf p}_{LK}
    \end{bmatrix}
\end{eqnarray}

where each ${\bf p}_{lk} \in \mathbb{R}^{D}$ denotes the proposal feature of the $l$-th scale in the $k$-th segment group, which is the average of the C3D features of the frames participating in the proposal.
Fully-connected layers are used to resize the feature dimension of $Q$ and $V$ to $D$. 
L2 normalization is performed to make each ${\bf p}_{lk}$ a unit vector.
%

\subsection{Surrogate Proposal Selection module}
To reduce the large number of proposals, \cite{SCN} proposed a sampling-based selection algorithm to prune out irrelevant proposals considering the exploration and exploitation. 
However, the method is trained using policy gradient algorithm \cite{PG} which suffers from high variance.
Instead, as depicted in Figure \ref{fig:proposal}(b), the Surrogate Proposal Selection module selects the best-matched proposals from each segment group based on the cosine similarity to the final hidden feature of the query.
A surrogate proposal of the $k$-th segment group is defined as the proposal that has the largest cosine similarity to the final hidden feature of the query.
The cosine similarity between each proposal and query is given by

\begin{eqnarray}
    \begin{bmatrix}
        {\bf p}_{11} \cdot {\bf w}_M & {\bf p}_{12} \cdot {\bf w}_M & \cdots & {\bf p}_{1K} \cdot {\bf w}_M \\
        {\bf p}_{21} \cdot {\bf w}_M & {\bf p}_{22} \cdot {\bf w}_M & \cdots & {\bf p}_{2K} \cdot {\bf w}_M \\
        \vdots &        & \cdots &        \\
        {\bf p}_{L1} \cdot {\bf w}_M & {\bf p}_{L2} \cdot {\bf w}_M & \cdots & {\bf p}_{LK} \cdot {\bf w}_M
    \end{bmatrix}
\end{eqnarray}
where ${\bf w}_M$ is the final hidden feature of the query. 
It is empirically determined that the final hidden query feature is sufficient in pruning out irrelevant proposals at a low computational cost.
The Surrogate Proposal Selection module pick the $l^{\prime}$-th scale from each $k$-th segment group which is given by,

\begin{eqnarray}
    & l^{\prime} = \mbox{argmax}\left[ {\bf p}_{1k} \cdot {\bf w}_M \ {\bf p}_{2k} \cdot {\bf w}_M \ \cdots \ {\bf p}_{Lk} \cdot {\bf w}_M \right], \\
    & {\bf s}_{k} = {\bf p}_{l^{\prime}k}
\end{eqnarray}
where ${\bf s}_k$ is the surrogate proposal feature of the $k$-th segment group. 
In backpropagation, only the surrogate proposals ${\bf s}_{k}$'s contribute to the weight update which allows end-to-end learning. 
Then the video is represented by $K$ surrogate proposal features:

\begin{eqnarray}
    \mathcal{V} = \left[ {\bf s}_1 \ {\bf s}_2 \ \cdots \ {\bf s}_K \right]
\end{eqnarray}

where $\mathcal{V}$ is the updated video representation composed of the surrogate proposals.

\subsection{Cascaded Cross-modal Attention module}
Cascaded Cross-modal Attention (CCA) module takes the video and query representations as inputs, and outputs a compact attended video representation. Compared to text-guided attention (TGA) \cite{that}, CCA module considers more diverse multi-modal feature interactions including V2V, Q2Q, V2Q, and Q2V where each has its own advantages as described below. 

\subsubsection{Dense Attention}
\label{sssec:dense_attention}
The basic attention unit of CCA module is referred to as Dense Attention which calculates the attention between two multi-element features. 
Given $Y=[{\bf y}_1 \ldots {\bf y}_M]^T\in \mathbb{R}^{M \times D}$ and $X=[{\bf x}_1 \ldots {\bf x}_N]^T \in \mathbb{R}^{N \times D}$, the Dense Attention $A(X,Y) : \mathbb{R}^{N \times D} \times \mathbb{R}^{M \times D} \rightarrow \mathbb{R}^{N \times D} $ attends $X$ using $Y$ and is defined as follows:

\begin{eqnarray}
& \mathcal{E}({\bf x}_n,Y) = \sum_{m=1}^{M} \mbox{tanh}(W_1 {\bf x}_n \cdot W_2 {\bf y}_m), \\
& A(X,Y) = \mbox{Softmax}(\left[ \mathcal{E}({\bf x}_1,Y) \ \mathcal{E}({\bf x}_2,Y) \ \cdots \ \mathcal{E}({\bf x}_N,Y)\right]) X,
\end{eqnarray}

where $W_1, W_2$ are learnable parameters. Here, $\mathcal{E}: \mathbb{R}^D \times \mathbb{R}^{M \times D} \rightarrow \mathbb{R}$ is referred to as the Video-Language Alignment (VLA) function that performs the multi-modal alignment.  

\subsubsection{Self-attention} 
Based on the Dense Attention defined above, the CCA module initially performs a type of self-attention that attends $\mathcal{V}$ and $Q$ using $\mathcal{V}$ and $Q$ respectively as given below,
\begin{eqnarray}
    \mathcal{V} \leftarrow A(\mathcal{V},\mathcal{V}), \\
    Q \leftarrow A(Q,Q).
\end{eqnarray}
The intra-attention allows each element of itself to be attended by its global contextual information. 
The attention from $\mathcal{V}$ to $\mathcal{V}$ is capable of highlighting the salient proposals by considering the innate temporal relationships. 
The attention from $Q$ to $Q$ updates the each word-level feature by considering the context of the whole sentence. 

\subsubsection{Cross modal attention} 
Following self-attention defined above, the CCA module is used to cross-attend $\mathcal{V}$ and $Q$ using $Q$ and $\mathcal{V}$ respectively such that cross-modal attention is defined as follows:
\begin{eqnarray}
    \mathcal{V} \leftarrow A(\mathcal{V},Q), \\
    Q \leftarrow A(Q,\mathcal{V}).
\end{eqnarray}
The above attention is critical in learning the latent multi-modal alignment. 
It has been empirically observed that cross-modal attention applied in series several times until near-saturation can be conducive in producing better performance.
Finally, a compact attended video representation ${\bf v}_{comp}$ is obtained by taking the sum of all elements of $\mathcal{V}$, and video-level similarity $c$ is obtained by the VLA function between ${\bf v}_{comp}$ and $Q$ as given below:

\begin{eqnarray}
    c = \mathcal{E}({\bf v}_{comp},Q).
\end{eqnarray}

The network is trained using the following contrastive loss:

\begin{eqnarray}
    \mathcal{L}_{contrastive} = max[0, \Delta - \mathcal{E}({\bf v}_{comp},Q^+) + \mathcal{E}({\bf v}_{comp},Q^-)]
\end{eqnarray}

where $\mathcal{E}$ is the VLA function defined above in section \ref{sssec:dense_attention} and $\Delta$ is the margin. $Q^+$ and $Q^-$ is positive and negative query features. 

\section{Experiment}
\subsection{Datasets}
\subsubsection{Charades-STA}
The Charades dataset was originally introduced in \cite{Charades}. 
It contains temporal activity annotation and multiple video-level descriptions for each video.
Gao \textit{et al.} \cite{TALL} generated temporal boundary annotations for sentences using a semi-automatic way and released the Charades-STA dataset that is for video moment retrieval.
The dataset includes 12,408 video-sentence pairs with temporal boundary annotations for training and 3,720 for testing. 
The average length of the query is 8.6 words, and the average duration of the video is 29.8 seconds.

\subsubsection{DiDeMo}
The Distinct Describable Moments (DiDeMo) dataset \cite{MCN} consists of over 10,000 unedited, personal videos in diverse visual settings with pairs of localized video segments and referring expressions. 
The videos are collected from Flickr and each video is trimmed to a maximum of 30 seconds. 
The dataset includes 8,395, 1,065 and 1,004 videos for train, validation, and test, respectively. 
The videos are divided into 5-second segments to reduce the complexity of annotation, which results in 21 possible moments per video. 
The dataset contains a total of 26,892 moments with over 40,000 text descriptions. 
The descriptions in the DiDeMo dataset are natural language sentences that contain activities, camera movement, and temporal transition indicators. 
Moreover, the descriptions in DiDeMo are verified to refer to a single moment.

\subsubsection{Evaluation Metric}
For Charades-STA, the evaluation metric proposed by \cite{TALL} is adopted to compute “R@n, IoU=m”. 
For the test set predictions, the recall R@n calculates the percentage of samples for which the correct result resides in the top-n retrievals to the query. 
If the IoU between the prediction and the ground truth is greater than or equal to $m$, the prediction is correct. 
The overall performance is the average recall on the whole test set. 

For DiDeMo, the evaluation metric proposed by \cite{MCN} is adopted. 
The evaluation metric is also R@n with different criteria for correct prediction. 
If the ground truth moment is in the top-n predictions, the prediction for the sample is counted as correct. 
The mIoU metric is computed by taking the average of the IoU between the predicted moment and the ground truth moment. 

\begin{table}[t]
	\centering
	\caption{Performance comparison of VLANet to the related methods on Charades-STA}\smallskip
	\begin{tabular}{c|l||c c c | c c c}
		\Xhline{3\arrayrulewidth}
		\multirow{2}{*}{Type}     & \multirow{2}{*}{Method} &  & R@1 &  &  & R@5 & \\ & & IoU=0.3 & IoU=0.5 & IoU=0.7 & IoU=0.3 & IoU=0.5 & IoU=0.7 \\ 		\Xhline{3\arrayrulewidth}
		Baseline & Random      & 19.78 & 11.96 & 4.81  & 73.62 & 52.79 & 21.53 \\ \hline
		\multirow{5}{*}{Fully} 
		         & VSA-RNN \cite{TALL}        &  -    & 10.50 & 4.32  & -     & 48.43 & 20.21 \\
		         & VSA-STV \cite{TALL}        & -     & 16.91 & 5.81  & -     & 53.89 & 23.58 \\ 
		         & CTRL \cite{TALL}           & -     & 23.63 & 8.89  & -     & 58.92 & 29.52 \\ 
		         & EFRC \cite{EFRC}           & 53.00 & 33.80 & 15.00 & 94.60 & 77.30 & 43.90 \\
		         & MAN \cite{MAN}             & -     & 46.53 & 22.72 & -     & 86.23 & 53.72
		\\ \hline
		\multirow{3}{*}{Weakly}
		         & TGA \cite{that}            & 32.14 & 19.94 & 8.84  & 86.58 & 65.52 & 33.51 \\
		         & SCN \cite{SCN}             & 42.96 & 23.58 & 9.97  & 95.56 & 71.80 & {\bf 38.87} \\
		         & \bf{VLANet (ours)}     & \bf{45.24} & \bf{31.83} & \bf{14.17} & \bf{95.70} & \bf{82.85} & 33.09 \\
		\Xhline{3\arrayrulewidth}
	\end{tabular}
	\label{tab:quantitative_1}
\end{table}

\begin{table}[t]
	\centering
	\caption{Performance comparison of VLANet to the related methods on DiDeMo}\smallskip
	\begin{tabular}{c|l||c c c}
		\Xhline{3\arrayrulewidth}
		Type & Method & R@1 & R@5 & mIoU \\ 		
		\Xhline{3\arrayrulewidth}
		\multirow{3}{*}{Baseline}
		& Upper Bound        & 74.75 & 100   & 96.05 \\ 
		& Random             & 3.75  & 22.50 & 22.64 \\
		& LSTM-RGB-Local \cite{MCN} & 13.10 & 44.82 & 25.13 \\ \hline 
		\multirow{5}{*}{Fully}
		& Txt-Obj-Retrieval \cite{TOR} & 16.20 & 43.94 & 27.18 \\
		& EFRC \cite{EFRC}   & 13.23 & 46.98 & 27.57 \\
		& CCA \cite{CCA}               & 18.11 & 52.11 & 37.82  \\ 
		& MCN \cite{MCN}               & 28.10 & 78.21 & 41.08  \\
		& MAN \cite{MAN}     & 27.02 & 81.70 & 41.16 \\ \hline
		\multirow{2}{*}{Weakly}
		& TGA \cite{that}    & 12.19 & 39.74 & 24.92 \\
		& {\bf VLANet (ours)}& {\bf 19.32} & {\bf 65.68} & {\bf 25.33}  \\ \hline
		
		\Xhline{3\arrayrulewidth}
	\end{tabular}
	\label{tab:quantitative_2}
\end{table}

\subsection{Quantitative result}
Table \ref{tab:quantitative_1} shows the performance comparison between VLANet and the related methods on Charades-STA. 
The first section indicates random baseline, the second section indicates fully-supervised methods, and the third section indicates weakly-supervised methods. 
VLANet achieves state-of-the-art performance on Charades-STA among weakly-supervised methods. 
It outperforms the random baseline, VSA-RNN, and VSA-STV by a large margin. 
Compared to the other fully-supervised methods such as CTRL and EFRC, its performance is comparable. 
Besides, compared to the other weakly-supervised methods TGA and SCN, VLANet outperforms others by a large margin. 

Table \ref{tab:quantitative_2} shows the performance comparison on DiDeMo.
The first section contains the baselines, the second section contains fully-supervised methods, and the third section contains weakly-supervised methods. 
VLANet achieves state-of-the-art performance among the weakly-supervised methods.
In the R@5 based test, especially, its performance is 25.94 higher than the runner-up model TGA. 
It is comparable to some fully-supervised methods such as CCA \footnote{Here, CCA refers to a previous method \cite{CCA}, but not Cascaded Cross-modal Attention proposed in this paper.} and Txt-Obj-Retrieval. 
These indicate that even without the full annotations of temporal boundary, VLANet has the potential to learn latent multi-modal alignment between video and query, and to localizing semantically relevant moments. 

\begin{table}[t]
	\centering
	\caption{Performance of model variants and ablation study of VLANet on Charades-STA. The unit of stride and window size is frame. }\smallskip
	\begin{tabular}{l||c c c | c c c}
		\Xhline{3\arrayrulewidth}
		\multirow{2}{*}{Method}      &  & R@1 &  &  & R@5 & \\ & IoU=0.3 & IoU=0.5 & IoU=0.7 & IoU=0.3 & IoU=0.5 & IoU=0.7 \\ 		\Xhline{3\arrayrulewidth}
		stride 4, window(176, 208, 240) & 44.76 & 31.53 & 14.78 & 77.04 & 63.17 & 31.80 \\
		stride 6, window(176, 208, 240) & 42.17 & 28.60 & 12.98 & 88.76 & 74.91 & {\bf 34.70} \\
		stride 8, window(176, 208, 240) & 45.03 & 31.82 & 14.19 & 95.72 & 82.82 & 33.33 \\
		stride 6, window(128, 256)     & 42.39 & 28.03 & 13.09 & 94.70 & 73.06 & 30.69 \\
		stride 6, window(176, 240)     & 42.92 & 30.24 & 13.57 & 95.72 & 82.80 & 33.46 \\ \hline
		
		w/o cross-attn     & 43.41 & 30.08 & 13.23 & 95.72 & 82.41 & 33.06 \\
		w/o self-attn   & 42.31 & 30.81 & 15.38 & 95.38 & 80.02 & 33.76 \\
		w/o surrogate   & 35.81 & 25.30 & 12.26 & 80.61 & 64.57 & 31.31 \\ \hline
		\bf{full model} & \bf{45.03} & \bf{31.82} & \bf{14.19} & \bf{95.72} & \bf{82.82} & 33.33 \\
		\Xhline{3\arrayrulewidth}
	\end{tabular}
	\label{tab:ablation}
\end{table}

\subsection{Model variants and ablation study}
Table \ref{tab:ablation} summarizes the performance of model variants and the ablation study conducted on VLANet.
The first section shows the performance variation by varying stride and window sizes, and the second section shows the performance drop without core components. 
The strides and the sizes of the windows were determined by considering the average video length. 
The first three rows show that the network performs best with the stride of 8.
While the proposals with stride 4 have finer granularity, the large number of proposals decreases the performance. 
The networks with three multi-scale proposals tend to achieve higher performance than the networks with two multi-scale proposals. 
This shows the importance of stride and the number of scales. 
After finding the best hyper-parameters of `stride 8, window(176, 208, 240)'\, these values were fixed for the subsequent experiments and analyses. 
The network without cross-attention, self-attention show a decrease in performance, demonstrating the importance of the attention mechanisms.
We generally notice a drop in performance with an increasing IoU metric. The drop is more drastic without cross-attention than without self-attention. This observation indicates that  cross-modal attention has a larger influence on performance than self-attention. 
The performance of w/o surrogate is decreased significantly across all metrics. This indicates that selecting probable proposals in the early stage is critical to the performance.

\subsection{Analysis of multi-modal similarity}
\begin{figure}[t]
	\centering
	\includegraphics[width=7cm]{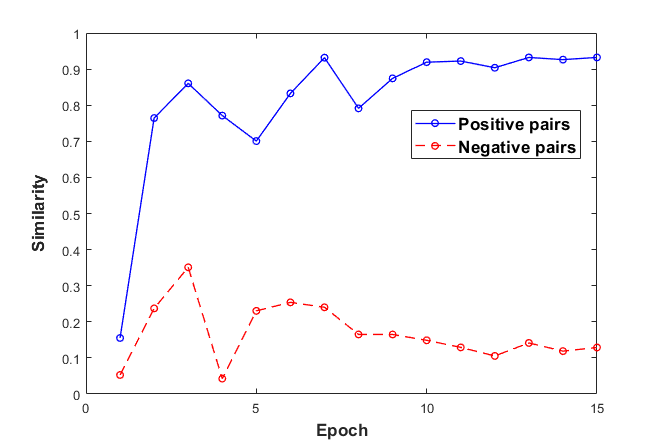}
	\caption{The multi-modal similarity prediction by VLANet on the positive and negative pairs while training. The similarity gap increases as epoch increases. }
	\label{fig:separation}
\end{figure}

Figure \ref{fig:separation} shows similarity predicted by the network on the whole test set of Charades-STA while training. 
The x-axis indicates the epoch of training, and the y-axis indicates the similarity. 
It is observed that the similarity scores of the positive pairs (blue) increase and reach a high plateau of about 0.9, while those of the negative pairs (red) keep a low value of about 0.15. 
These demonstrate that contrastive learning was successfully conducted. 

\begin{figure}[t]
	\centering
	\includegraphics[width=\textwidth]{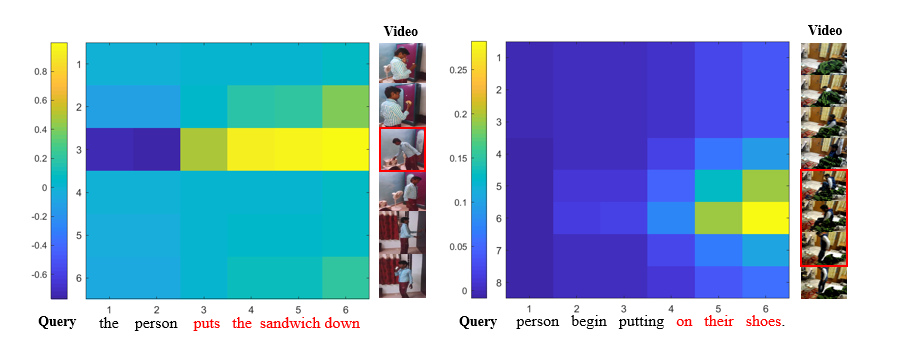}
	\caption{Visualization of Cascaded Cross-modal Attention. The attention map is calculated by the outer-product of video and query features that are obtained after the Cascaded Cross-modal Attention module and before the 
	pooling layer. }
	\label{fig:attention}
\end{figure}

\subsection{Visualization of attention map}
Figure \ref{fig:attention} visualizes the attention map of the proposed Cascaded Cross-modal Attention. 
The x-axis indicates the words in the query and the y-axis indicates the time. 
In the left example, the attention weight of the ``put the sandwich down" is high when the person is putting the sandwich down.
Similarly in the right example, important words such as action or object have high attention weight with the related moment of the video. 
The high attention weights are biased on the right side in Figure \ref{fig:attention} as the final GRU feature has the context information about the whole sentence. 
The above example demonstrates that VLANet can learn the latent multi-modal alignment. 

\begin{figure}[t]
	\centering
	\includegraphics[width=11cm]{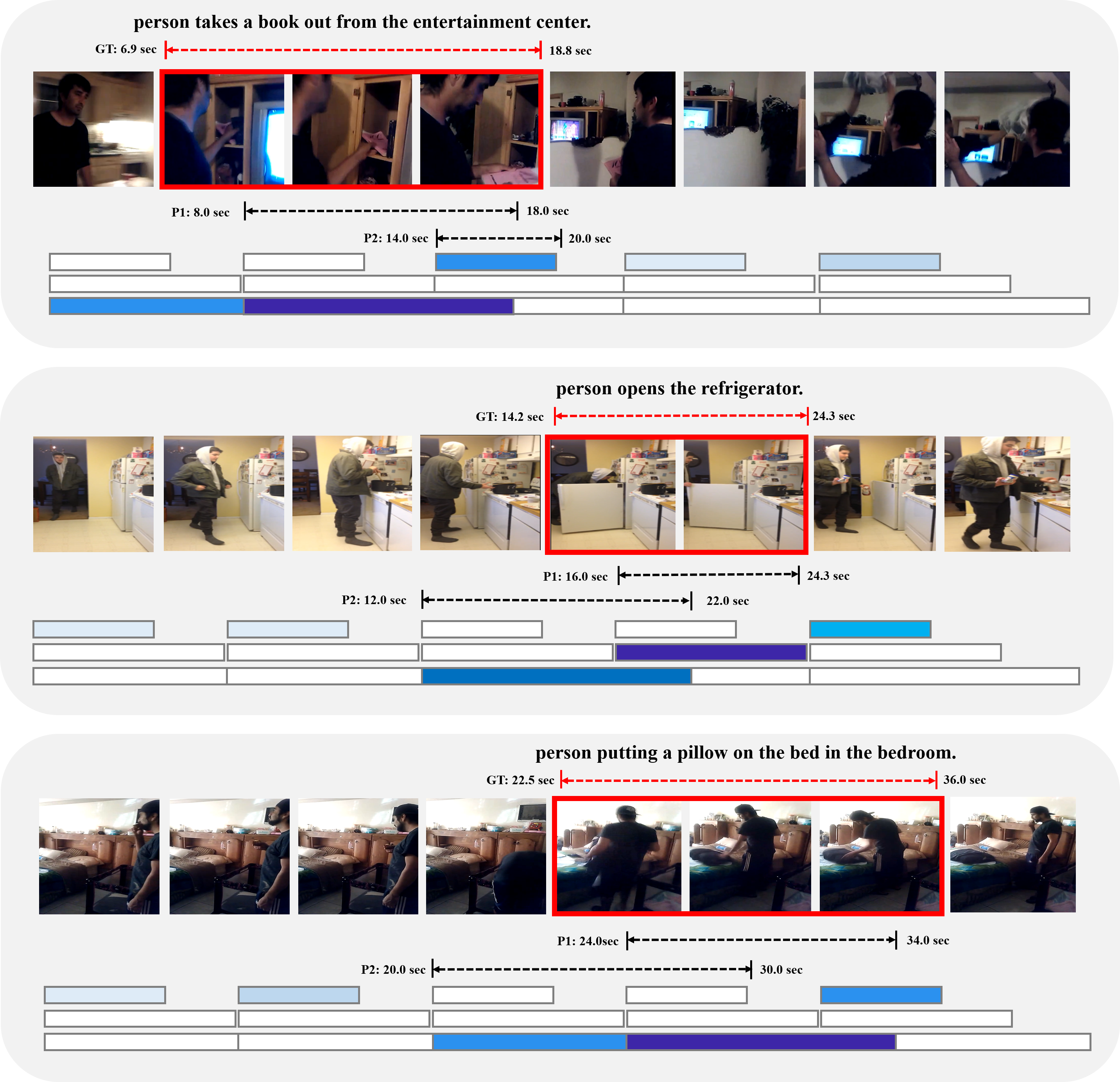}
	\caption{At inference time, VLANet successfully retrieves the moment described by the query. Due to the limited space, only some proposals are visualized. The color indicates the attention strength. The top-2 predicted moments are visualized with the temporal boundaries. }
	\label{fig:inference}
\end{figure}

\subsection{Visualization of inference}
Figure \ref{fig:inference} provides a visualization of the inference of VLANet. 
Only a subset of total proposals were depicted whose color indicates the attention strength. 
In the first example, both top-1 and top-2 predictions by VLANet have high overlaps with the ground truth moment. 
In the second example, the network localizes the moment when the person actually \textit{opens} the refrigerator.
Similarly in the third example, the network localizes the moment when person \textit{puts} the pillow. 
This shows that the network successfully captures the moment when a certain action is taken or an event occurs. 
The inference visualization demonstrates the moment retrieval ability of VLANet and suggests its applicability to real-world scenarios.

\section{Conclusions}
This paper considers Video-Language Alignment Network (VLANet) for weakly-supervised video moment retrieval. 
VLANet is able to select appropriate candidate proposals using a more detailed query representation that include intermediate hidden features of the GRU.
The Surrogate Proposal Selection module reduces the number of candidate proposals based on the similarity between each proposal and the query. The ablation study reveals that it has the largest influence on performance. 
The Cascaded Cross-modal Attention module performs a modified self-attention followed by a cascade of cross-attention based on the Dense Attention defined. It also has a significant influence on performance.
VLANet is trained in an end-to-end manner using contrastive loss which enforces semantically similar videos and queries to cluster in the joint embedding space. 
The experiments shows that VLANet achieves state-of-the-art performance on Charades-STA and DiDeMo datasets. 

\section*{Acknowledgement}
This work was partly supported by Institute for Information \& communications Technology Planning \& Evaluation(IITP) grant funded by the Korea government(MSIT) (2017-0-01780, The technology development for event recognition/relational reasoning and learning knowledge based system for video understanding) and partly supported by Institute for Information \& communications Technology Planning \& Evaluation(IITP) grant funded by the Korea government(MSIT) (No. 2019-0-01396, Development of framework for analyzing, detecting, mitigating of bias in AI model and training data)

\clearpage
%
%
\bibliographystyle{splncs04}
\bibliography{egbib}
\end{document}